# Utility-Based Abstraction and Categorization

Eric J. Horvitz* and Adrian C. Klein
Palo Alto Laboratory
Rockwell International Science Center
444 High Street
Palo Alto, CA 94301

## Abstract

We take a utility-based approach to categorization. We construct generalizations about events and actions by considering losses associated with failing to distinguish among detailed distinctions in a decision model. The utility-based methods transform detailed states of the world into more abstract categories comprised of disjunctions of the states. We show how we can cluster distinctions into groups of distinctions at progressively higher levels of abstraction, and describe rules for decision making with the abstractions. The techniques introduce a utility-based perspective on the nature of concepts, and provide a means of simplifying decision models used in automated reasoning systems. We demonstrate the techniques by describing the capabilities and output of TUBA, a program for utility-based abstraction.

## 1   INTRODUCTION

There has been long-term interest in cognitive and computational models for transforming a set of detailed attributes or concepts into more general concepts. Most methods employed to date for categorization are based on a consideration of similarities in the attributes of different objects (Rosch and Lloyd, 1978; Smith and Medin, 1981; Schank et al., 1986; Fisher, 1987; Ashby and Gott, 1988; Medin, 1989). We take a decision-analytic perspective on the generation of categories and concepts by considering losses associated with the clustering of distinctions about events and actions. By tolerating increasing imprecision in the utilities associated with the outcomes of actions, we can generate increasingly abstract categories of states of the world and of actions. The methods can be applied at design time, or in real time, for reducing the size, and, potentially the computational complexity of belief networks and influence diagrams.

In earlier work, we explored the simplification of computational models of decision making through generalizing the distinctions considered in a decision model (Horvitz et al., 1989). In that work, we increased the speed of computation and the ease of explanation of the results of automated medical diagnosis by employing abstraction hierarchies defined by expert physicians to group diseases into categories of disease. In this paper, we explore methods for automating the construction of abstractions, and hierarchies of abstractions, based on utility considerations.

We start by considering ideal actions under uncertainty, given a detailed utility model, and show how we can generalize the approach to consider groups consisting of disjunctions of events. We describe some empirical studies of utility-based abstraction using TUBA, a program for utility-based abstraction. Then, we discuss decision making with abstract categories.

## 2   ACTIONS UNDER UNCERTAINTY

The expected value of an action depends on the likelihoods of different states of the world, or events, and on the possible outcomes that follow from that action. Assume that there are $H_1, \ldots, H_n$ mutually exclusive and exhaustive states of the world. A decision makers action $A_i$, taken in the context of a state of the world $H_i$, defines an outcome $(A_i, H_j)$. We use $u(A_i, H_j)$ to refer to the utility of a decision maker who takes an action (or set of actions) $A_i$ when state $H_j$ is true. The value of different actions under uncertainty depends on the probability of different events, and the result or outcome of different actions, given these probabilities. Assume that a decision making agent has gathered a set of evidence $\mathbf{E}$ about its environment (e.g., sensors or direct observations), and employs probabilistic inference over a belief-network to compute a probability distribution over a set of mutually exclusive and exhaustive hypotheses, $p(H|\mathbf{E}, \xi)$, where $\xi$ represents the background state of information. Given such a distribution, the expected utility, eu, of each action $A_i$

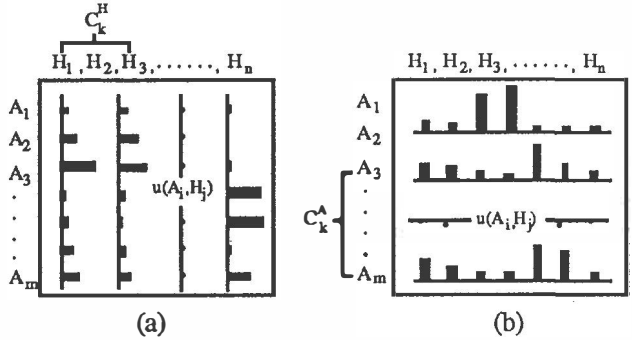

Figure 1: Given a utility model for a set of actions $A_1$ through $A_m$ and states of the world $H_1$ through $H_m$, we wish to group hypotheses into categories of hypotheses ($C_k^H$) (a) or into categories of actions ($C_k^A$) (b) by considering the similarity of utilities of outcomes, $u(A_i, H_j)$ (bar-graph heights).

is

$$\text{eu}(A_i) = \sum_{j=1}^{n} p(H_j|\mathbf{E}, \xi) u(A_i, H_j) \quad (1)$$

and the ideal decision, $A^*$, is the action with the greatest expected utility, given the probability distribution and the set of utilities,

$$A^* = \arg\max_{A_i} \sum_{j=1}^{n} p(H_j|\mathbf{E}, \xi) u(A_i, H_j) \quad (2)$$

In Section 6, we explore decision making with abstract categories of events versus atomic events $H$. First, we consider the generation of categories by introducing tolerance for error in the utilities assigned to outcomes.

## 3 ABSTRACTION BY UTILITY-BASED SIMILARITY

Given a utility model, we can reduce the size of decision models, and thus, the computational or cognitive requirements of decision making, by generating abstract categories from base-level distinctions.

### 3.1 Categorization of World States

Assume that we have a set of mutually exclusive and exhaustive states of the world of interest, and wish to generate a set of disjoint categories. More specifically, we seek to identify categories $C^H$, defined as a set of base-level events $H$. We interpret a category of events as a *disjunction* of states of the world $C_k^H \rightarrow (H_1 \vee H_2 \vee, \ldots, \vee H_m)$, and consider the utility of actions given the probability of alternate categories.

We generate abstractions by grouping states of the world that are associated with a similar pattern of utilities, with repsect to a given set of feasible actions. We can construct groups by progressively increasing the number of terms in disjunctions of base-level hypotheses. To build categories, we employ a search algorithm in conjunction with a threshold on error in utilities for taking actions, given the possibility of any states in a category.

Let us focus on the construction of a set of disjoint categories of hypotheses $H$, based on a process of identifying similarities in the utility of outcomes. For each action $A \in \mathbf{A}$ of feasible actions, and each category $C^H$, we determine the maximum difference or *span* in utility associated with taking that action, when any one of several states $H \in C^H$ might be true. We identify the maximum and minimum utilities of taking action when $C_k^H$ is true, and compute the difference, or *span* in utility represented by the outcomes associated with a group of states. That is,

$$\text{Uspan}^H = \max_{H_j \in C_k^H} u(A_i, H_j) - \min_{H_j \in C_k^H} u(A_i, H_j) \quad (3)$$

where $\text{Uspan}^H(A_i, C_k^H)$ is the breadth of the range of utility values associated with an outcome when considering an action in a context where we only know that the disjunction $C_k^H$ is true, as compared with decision making with detailed elements $H_j$. The maximum span of utility encountered with the consideration of the presence of a group, versus an explicit consideration of outcomes in terms of each of its disjuncts is just the maximum of all the Uspan measures,

$$\max \text{Uspan}^H = \max_{A_i \in \mathbf{A}} \text{Uspan}^H \quad (4)$$

We can employ preferences about maxUspan to control the size, and thus, the number of categories used in decision making.

Employing a general search to identify all appropriate abstractions is computationally intractable. In Section 4, we describe polynomial clustering methods to build a hierarchy of categories and to select a level of abstraction at which the range of utility values is always less than a specified maxUspan tolerance. In Section 6, we discuss decision making with abstract disjunctions in lieu of base-level hypotheses. First we review an analogous approach for categorizing actions.

### 3.2 Categorization of Actions

We generate abstractions about *actions* in a manner analogous to the way we generate abstractions of *states* of the world. Rather than group states, we group actions that are associated with a similar pattern of utilities, as we consider all feasible hypotheses. We identify categories of actions $C^A$ by identifying groups of actions associated with outcomes that have utilities within an acceptable range, with respect to states of the world $H \in \mathbf{H}$. As decision making agents can take only a single action at any time, we interpret a decision to commit to a specific category of actions as taking one of any actions $A \in C^A$ in a group.

For each category $C^A$ and each hypothesis $H \in \mathbf{H}$ of events of interest, we determine the span in utility gen-erated as we assume different states of the world. We identify the maximum and minimum utilities of taking any action from $C_k^A$, when $H_j$ is true, and compute the difference

$$\text{Uspan}^A = \max_{A_i \in C_k^A} u(A_i, H_j) - \min_{A_i \in C_k^A} u(A_i, H_j) \quad (5)$$

$\text{Uspan}^A$ is the span of utility associated with taking one of any actions from group $C^A$ when $H_j$ is true. The maximum range in utility associated with use of groups of actions versus an explicit consideration of detailed outcomes is,

$$\text{maxUspan}^A = \max_{H_j \in \mathbf{H}} \text{Uspan}^A(C_k^A, H_j) \quad (6)$$

We can consider the *expected span* in utility, $\text{EUspan}^A(C_k^A, H_j)$, for groups of actions, and the maximum of this expectation, by weighting the span, associated with each category of actions and state, by the probability of each state. That is,

$$\text{EUspan}^A =$$

$$p(H_j|\mathbf{E}, \xi) \left[ \max_{A_i \in C_k^A} u(A_i, H_j) - \min_{A_i \in C_k^A} u(A_i, H_j) \right] \quad (7)$$

We can assume for the probability of each state, posterior probabilities computed explicitly, or, assume as a heuristic, prior probabilities of hypotheses. Alternatively, a system engineer may wish to encode distinct sets of categories in terms of contexts defined by common patterns of evidence.

## 4 POLYNOMIAL COMPUTATION OF ABSTRACTIONS

As highlighted in Figures 1(a) and 1(b), we seek a tractable means of identifying hypotheses that are similar in terms of the utilities of the set of outcomes generated by crossing the hypotheses with a set of actions, and in analogous analyses to generate categories of actions in terms of the similarity in utility of outcomes across sets of events. We focus in this section on pragmatic concerns with regard to utility-based grouping of events and actions, in accordance with a maximal allowed span in utility for categories. We have experimented with several polynomial algorithms for building clusters of events based on the utility of outcomes, so as to identify categories and hierarchies of categories containing outcomes at increasingly greater differences in utility. Building hierarchies of categories at increasing levels of abstraction, allows us to generate sets of categories with different maximal spans in utility.

Several practical utility-based categorization methods, and auxiliary abstraction facilities, are embodied in a program named TUBA. The program runs on the Apple Macintosh family of computers. TUBA takes as input a utility model and outputs an abstraction hierarchy of categories based on similarities in the utility of outcomes.

### 4.1 Distances and Similarity in Utility Space

TUBA constructs categories by clustering of hypotheses by similarity in outcome utility. As portrayed in Figure 2, the task of generating utility-based abstractions can be viewed as the delineation of boundaries around clusters of events in a geometric collection of points representing hypotheses or actions in an $n$-dimensional utility space. Several distance metrics can be used to cluster hypotheses based on preferences about losses associated with generalization. We can cluster atomic events into clusters of events with a goal of minimizing the maximum range of utility values associated with outcomes of decisions based on a consideration of the likelihood of categories. This can be accomplished by using the maximum Uspan as a metric to drive such clustering. Alternatively, we can build concepts that capture an intention to minimize *expected* losses of decisions with abstract concepts. For such clustering, we categorize events and actions based on minimizing the Euclidean distance in a utility space. We can employ an unweighted Euclidean distance, or a distance metric that is weighted to take into consideration the different probabilities of events.

With a Euclidean distance metric, we compute the distance between vectors of utilities of outcomes in $n$ dimensions, reflecting each of $n$ actions under consideration. For building categories of events, we compute $D$ for any two hypotheses, $H_1$ and $H_2$, as,

$$D(H_1, H_2) = \sqrt{\sum_{i=1}^{n} [u(A_i, H_1) - u(A_i, H_2)]^2} \quad (8)$$

For building categories of actions, we compute $D$ for any two actions, $A_1$ and $A_2$, as,

$$D(A_1, A_2) = \sqrt{\sum_{i=1}^{n} [u(A_1, H_i) - u(A_2, H_i)]^2} \quad (9)$$

For building categories of actions, based on a metric of expected distance, we compute $D$ for any two actions, $A_1$ and $A_2$, $D(A_1, A_2) =$

$$\sqrt{\sum_{i=1}^{n} (p(H_i|\mathbf{E}, \xi) [u(A_1, H_i) - u(A_2, H_i)])^2} \quad (10)$$

so that differences in utility of actions, given the occurence of world states, are weighted according to the probability of the states. With application of this distance metric, categories can be constructed for prior probability distributions, or can be dynamically reformulated given changes in the posterior probabilities of world states as evidence is observed.

### 4.2 Utility-Based Abstraction Hierarchies

We have examined several different utility-based clustering algorithms for building hierarchies of categories.

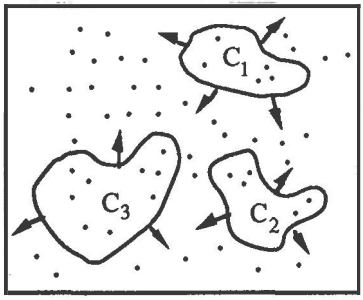

Figure 2: We construct a hierarchy of categories by clustering outcomes by similarity in a multidimensional utility-space.

The methods for building abstractions available in TUBA are adaptations of traditional clustering methods (Johnson and Wichern, 1982). The methods differ as to how distance between two groups of hypotheses is defined. Both methods start with the set of base-level, atomic hypotheses as groups. The two closest hypotheses are merged and distances between all groups are updated to reflect the merger. The merger procedure continues until all hypotheses or actions have been merged into a single group. At each merger, the distance between the two groups being merged is recorded. The *complete-linkage* method defines the distance between two groups as the greatest distance between any member of one group and any member of the other. For hypothesis clustering, that is

$$D(C_1, C_2) = \max [D(H_i, H_j)] \quad (11)$$

where $H_i \in C_1$ and $H_j \in C_2$.

The *single-linkage* method, in contrast, takes the distance between two groups to be defined by their closest members. In practice, the complete-linkage method is generally preferable, since at each stage it minimizes the maximum cost of error based in failing to distinguish among members of the same group.

The result of hierarchical clustering based on utility is summarized graphically by an abstraction hierarchy of categories, with atomic events as leaves. Vertical lines extend upward from each group, and a horizontal line joining two vertical lines indicates a merger. The height of line indicating a merger indicates the distance between the two groups being merged.

If Uspan is used as a distance metric, the level of the hierarchy is the maximum span of utility of the groups formed by a merger. We can select a maximal level of abstraction by noting the level at which categories exceed a preferred maximum span of utility. Categories that lie just below this line are admitted; disjunctions of states formed by mergers above this cutoff represent groups in which the maximum span has been exceeded. Similarly, when complete-linkage is used, we can specify a cutoff in terms of maximum distance in $n$-space. The two methods are closely linked; whereas complete-linkage can be viewed geometrically as mimimizing the span of a group across a Euclidean hyperspace, the use of the Uspan metric minimizes the span of a group across each individual axis in the same hyperspace.

### 4.3 Extensions of the General Approach

We have explored several extensions of the basic utility-based approach to constructing categories of events, including the use of multiattribute utility and considering subsets of actions and hypotheses. These facilities are available in the TUBA program.

**Abstraction for Multiattribute Utility.** Preferences about outcomes may be represented as a function of several independent variables (Keeney and Raiffa, 1976; Keeney, 1977). For example, decision analysts often represent preference with an additive multiattribute model. TUBA allows the user to specify multiattribute utility models, and to explore how altering the weights of a utility model affect the classifications generated.

**Subsets of Actions and Hypotheses.** Rather than examining a distance vector of size defined by all available actions or hypotheses, we may wish to explore categories for subsets of actions or hypotheses. For example, in constructing categories in the context of a study on antibiotics, records containing detailed information about the response of diseases to therapy might be categorized solely on the basis of the utility of outcomes of antibiotic therapy, ignoring the outcomes associated with other therapy actions. TUBA allows users to specify arbitrary sets of actions, to allow for the generation of utility-based abstractions for different categories.

## 5  EXAMPLES OF UTILITY-BASED ABSTRACTION

We shall review examples of utility-based abstractions for robot decision making and medical diagnosis created by TUBA.

### 5.1  Robot Decision Making

Consider the problem domain of an autonomous robot developed to roam the corridors of a computer-science department in search of trash. The robot has the ability to perform four basic actions: (1) locate a socket to recharge its batteries while scanning an area for garbage, (2) meander about and record the location of trash, (3) actively gather refuse into its trashbag, and (4) beep to request assistance about the location of garbage. Engineers are faced with the task of developing visual sensors and a belief network to generate probabilities about the location of the robot. The robot's engineers initially divide the department into a set of six types of location: (1) a hallway, (2) a closet, (3) a restroom, (4) a stairwell, (5) a classroom, and (6) an office. They assess a utility model for the 24 possible outcomes $u(A_i, H_j)$. The designers wish to maximize the rate at which the robot collects garbage but minimize the annoyance of robot operation to people at the department. They develop an additive multiattribute utility model which weights outcomes of actions in terms of the efficiency of garbage collection and the degree to which the robot operates without distracting or annoying research staff and students. They specify a multiattribute function, $U = Q(q) + R(r)$, where $r$ represents garbage collecting efficiency, and $q$ is the degree to which the robot is quiet and unobstructive. Table 1 is believed to accurately describe preferences of the design team. $Q$ and $R$ are initially assigned the values of 0.1 and 0.9, respectively.

| Action/Location | Hallway | | Closet | | Office | | Stairwell | | Restroom | | Class | |
|---|---|---|---|---|---|---|---|---|---|---|---|---|
| Charge/Scan | q=0.6 | r=0.2 | 1.0 | 0.8 | 0.9 | 0.6 | 0.7 | 0.1 | 0.9 | 0.1 | 0.8 | 0.5 |
| Query Assist | 0.8 | 0.8 | 0.9 | 0.1 | 0.4 | 0.8 | 0.8 | 0.4 | 0.2 | 0.7 | 0.1 | 0.8 |
| Meander/Scan | 0.6 | 0.5 | 0.9 | 0.3 | 0.7 | 0.7 | 0.3 | 0.1 | 0.7 | 0.7 | 0.4 | 0.6 |
| Gather | 0.5 | 0.7 | 1.0 | 0.8 | 0.4 | 0.6 | 0.3 | 0.2 | 0.3 | 0.3 | 0.4 | 0.5 |

Table 1: Multiple attributes of utility for a refuse-collecting robot.

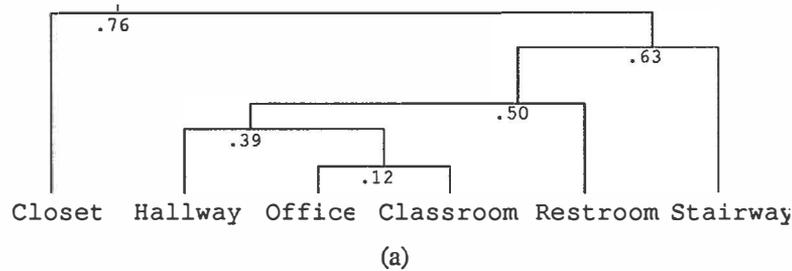

(a)

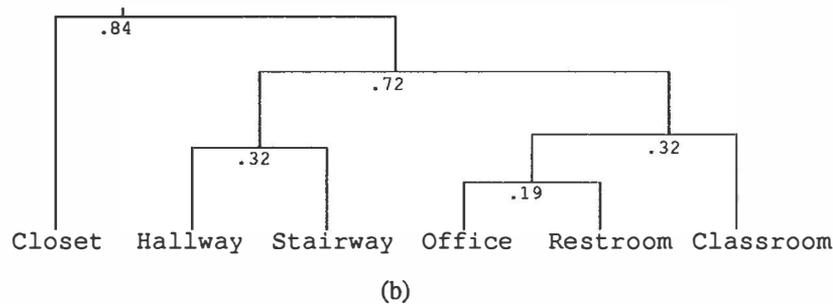

(b)

Figure 3: Utility-based abstraction hierarchies generated by TUBA for reducing the complexity of sensors and decision model of a wandering robot. Different sets of abstractions are generated by changing the weightings of a multiattribute utility function. The maximum span in utility of groups of events is indicated in labels at the merger lines.

Suppose that the robot's information about its location come from cues in its environment (e.g., preexisting items or custom-tailored color coding of baseboards of rooms in the department). The robot's designers wish to reduce the number of cues that the robot needs to distinguish, so as to simplify visual processing algorithms and reduce hardware requirements. Thus, the researchers apply TUBA to analyze the complete utility model of building areas. Using the initial multiattribute weightings, the optimal classifications, based on a utility-based Euclidean distance metric and the complete-linkage algorithm, are represented by the abstraction hierarchy displayed in Figure 3(a). The maximum possible loss of utility associated with the robot misclassifying its location among locations grouped into a category is printed at the merger line defining new groups. Based on this analysis, and a decision to tolerate a predefined error in the utility, the engineers decide to consider classrooms, offices, and hallways as a single group for the purposes of the robot's sensor discrimination and reasoning apparatus.

After several weeks of allowing the robot to roam through the department, the robotics group receives a note from the departmental administrator. Apparently, the robot has been disrupting several classes and important meetings in the building. To reduce the risk of department administration developing a policy restricting the robot's autonomous roaming, the engineers decide to consider a new utility model, and to redesign the reasoning system and sensor array. The revised utility model places more weight on the robot becoming less conspicuous, with $Q = 0.9$ and $R = 0.1$. These new coefficients result in a revised utility-based abstraction hierarchy displayed in Figure 3(b). The technicians now redesign the robot with sensors and uncertain reasoning apparatus for three classes of states, describing the location of the robot: office-restroom-classroom, stairway-hallway, and closet.

## 5.2 Medical Decision Making

We have applied utility-based abstraction to medical diagnosis and therapy problems to generate categories of therapy actions and disorders. Figure 5 displays TUBA output of an abstraction hierarchy of sets of actions generated from a detailed utility model for the diagnosis and treatment of lymph-node pathology. The utility model was developed and assessed for use in the Pathfinder pathology diagnostic system (Heckerman et al., 1992; Heckerman and Nathwani, 1992). The utility model represents preferences about 3600 outcomes. The model represents the utility of disease–treatment outcomes associated with a correct and erroneous diagnoses, where it is assumed that, should a disease be misdiagnosed and mistreated, the correct diagnosis will be made after some predefined length of time.

Utility-based categories of diseases in lymph-node pathology identified by TUBA using an unweighted Euclidean distance are displayed in Figure 4. Figure 5 demonstrates the identification of categories of therapy for lymph-node diseases by utility-based abstraction procedures. These classes of therapy include treatment for infection, Hodgkin's lymphomas, and non-Hodgkin's lymphomas. Note that, as one might expect, HIV is identified as an important distinguished entity in the disease categorization abstraction. However, it is admixed, at the same level as many other entities, in the *treat as infectious-benign* category in the treatment categorizations because the treatments for AIDS have relatively few side-effects and delays in treating many of the benign and infectious diseases affect patients minimally.

## 6 DECISIONS WITH ABSTRACTIONS

The utility-based construction of categories can be used solely as a means of posing to engineers valuable and simplifying generalizations about events and actions. Utilities of outcomes defined by generalizations of base-level distinctions can be assessed directly and the resulting abstract utility models can be used in automated decision-analytic reasoning. However, it is also possible to reason about utilities and decisions in terms of the base-level distinctions.

### 6.1 Decisions with Categories of Events

Let us first consider decision making based on a consideration of categories of states of the world, in lieu of atomic events. The probability of a disjunction of mutually exclusive events, $p(C_k^H|\mathbf{E},\xi)$, is the sum of the probabilities of its disjuncts,

$$p(C_k^H|\mathbf{E},\xi) = \sum_{H_i \in C_k^H} p(H_i|\mathbf{E},\xi) \qquad (12)$$

The expected utility of an action $A$, given the utilities assigned to actions when a disjunction is true, given a probability distribution over categories is,

$$\text{eu}(A_i) = \sum_{j=1}^{n} p(C_j^H|\mathbf{E},\xi)u(A_i,C_j) \qquad (13)$$

We select the decision $A^*$ with the maximum expected utility, given the probability distribution over categories $C$, as described in Equations 1 and 2.

What point utilities should we assign to abstract outcomes $u(A_i, C_j)$? If the uncertain-reasoning machinery is available we can compute directly the utility of taking an action given the truth of a category as,

$$\text{eu}(A_i, C_k^H) = \sum_{H_j \in C_k^H} p(H_j|C_k^H, \mathbf{E}, \xi)u(A_i, H_j) \qquad (14)$$

and can substitute the result of this calculation into Equation 14. However, given the theme of attempting to simplify multiple components of a decision model, the probabilities for each $H_j$ may not be available. If this is the case, we can employ expectations of actions for groups based on prior probabilities, or on prototypical contexts defined by common sets of evidence. A special case of computing the expected utility of action, given a group of events, is the case where we consider all hypotheses to be equally likely, given the truth of a category. The expected utility is equivalent to taking the average of the utilities, $\bar{u}(A_i, C_j^H)$,

$$\bar{u}(A_i, C_j^H) = \sum_{H_j \in C_k^H} \frac{u(A_i, H_j)}{||C_k^H||}$$

where $||C_k^H||$ is the cardinality of the set of hypotheses $C_k^H$.

Rather than making decisions based on expectation over utilities with a predefined tolerance of error, we can employ a minimax utility-bounding approach to decision making. We seek to determine whether the minimum expected utility associated with an action dominates the maximum expected value associated with all other actions. If this is true, we know that the leading action dominates the other actions, given error associated with abstraction. That is, we store only the minimum and maximum values of $u(A_i, C_j^H)$ and seek to identify $A^*$ that uniquely satisfies the following,

$$\sum_{j=1}^{n} p(C_j^H|\mathbf{E},\xi) \min_{H \in C_j^H} [u(A^*, H)] \geq$$

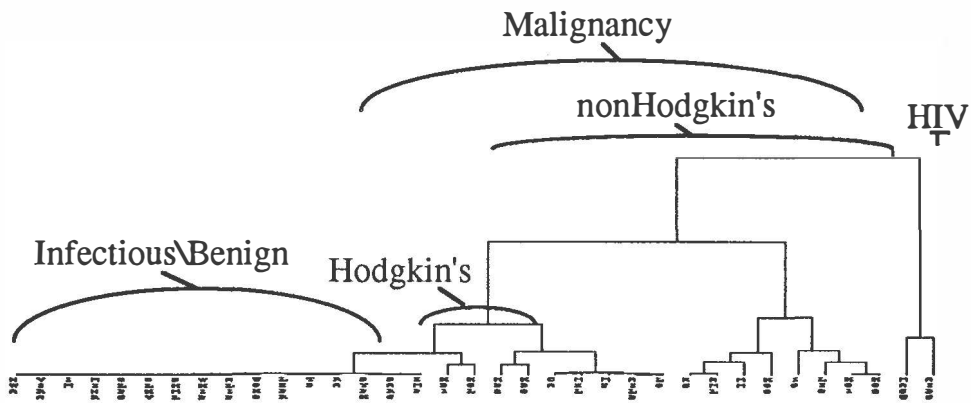

Figure 4: Disease categories generated by applying utility-based abstraction to a detailed utility model for oncology.

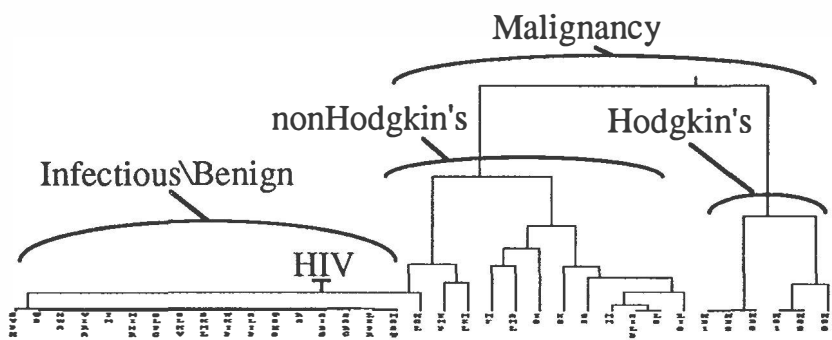

Figure 5: Therapy categories generated by applying utility-based abstraction to a detailed utility model for oncology.

$$\sum_{j=1}^{n} p(C_j^H|\mathbf{E},\xi) \max_{H \in C_j^H}[u(A,H)] \quad (15)$$

for $A \in \mathbf{A}, A^* \neq A$.

### 6.2 Decisions with Categories of Actions

Groups of actions in decision making differ from the consideration of groups of events in that a decision maker can take only a single action. One approach to simplifying decision models with the result of utility-based categorization of actions is to select a single base-level action $A$ from each category $C_k^A$, and to use these actions, and their associated utilities $u(A, H_j)$, to make decisions based on Equations 1 and 2. To minimize losses with considering a reduced set of actions, we can select, from each group of actions, that action with the highest expected utility, given an assumed probability distribution over events, $p(H|\mathbf{E},\xi)$. That is, we select $A^*$ for each $C_k^A$,

$$A^* = \arg\max_{A_i \in C_k^A} \sum_{H_j} p(H_j|\mathbf{E},\xi) u(A_i, H_j) \quad (16)$$

As we may not wish to continually compute these actions, based in a continually updated probability distribution, we may wish to preselect the set of actions based on the prior probability distribution over events, or on posterior probabilities for a set of contexts.

We can also employ a minimax bounding methodology to make decisions at the level of categories of action, analogous to the bounding method we described for making decisions with categories of states. We seek to determine whether the minimum expected utility associated with taking any action $A \in C_k^A$ dominates the maximum expected utility of taking any actions that are elements of other groups of actions. We store only the minimum and maximum values of $u(C_i^A, H_j)$ and search for an action $C^{A*}$ such that,

$$\sum_{j=1}^{n} p(H_j|\mathbf{E},\xi) \min_{A \in C^{A*}}[u(C^{A*}, H_j)] \geq$$

$$\sum_{j=1}^{n} p(H_j|\mathbf{E},\xi) \max_{A \in C^A}[u(C^A, H_j)] \quad (17)$$

for $C^A \in \mathbf{C^A}, C^{A*} \neq C^A$.

## 7 SUMMARY AND CONCLUSIONS

We described a utility-based approach to generating categories, and presented examples of the application of the methods in robotics and medical diagnosis. The utility-based methods complement the more familiar similarity-based and probability-based approaches to the construction and interpretation of concepts. We believe that many commonsense natural categories about events and actions have a basis in the similarity of the utility of outcomes. Utility-based categorization and abstraction can be useful in engineering decision systems, given constraints in modeling or computational resources. Beyond direct application of the abstraction methods to reduce detailed distinctions to categories, the hierarchical abstraction methods can offer experts and engineers intuitions about the level of detail at which to frame a decision problem.

Utility-based categorization methods also provide an additional tool for exploring rational decisions under bounded resources. In particular, the abstraction methods provide a means of trading off the complexity of reasoning with the precision of decision models. Beyond application of utility-based abstraction in the engineering of automated reasoning systems, the methods hold promise for dynamic, real-time application in agents that are forced to make decisions under varying and uncertain resource constraints (Horvitz, 1990). For example, when combined with an explicit model of the cost of reasoning as a function of the size of the action and outcome space, utility-based abstraction methods can be used to select the ideal level of detail at which to perform automated reasoning. We invite others to join us in experimenting with utility-based categorization; the TUBA program is available to interested researchers.

## References

Ashby, G. and Gott, R. (1988). Decision rules in the perception and categorization of multidimensional stimuli. *Journal of Experimental Psychology*, 14(1):33–53.

Fisher, D. H. (1987). Knowledge acquistion via incremential conceptual clustering. *Machine Learning*, 2:139–172. In J.W. Shavlik and T.G. Dietterich (eds.), *Readings in Machine Learning*, pp 267–283, Morgan Kaufmann, 1990.

Heckerman, D., Horvitz, E., and Nathwani, B. (1992). Toward normative expert systems: Part I. The Pathfinder project. *Methods of information in medicine*, 31:90–105.

Heckerman, D. and Nathwani, B. (1992). An evaluation of the diagnostic accuracy of Pathfinder. *Computers and Biomedical Research*, 25(1):56–74.

Horvitz, E. (1990). *Computation and Action Under Bounded Resources*. PhD thesis, Stanford University.

Horvitz, E., Heckerman, D., Ng, K., and Nathwani, B. (1989). Heuristic abstraction in the decision-theoretic Pathfinder system. In *Proceedings of the Thirteenth Symposium on Computer Applications in Medical Care, Washington, DC*, pages 178–182. IEEE Computer Society Press, Los Angeles, CA.

Johnson, R. and Wichern, D. (1982). *Applied Multivariate Statistical Analysis*. Prentice-Hall, Englewood Cliffs, NJ.

Keeney, R. (1977). The art of assessing multi-attribute utility functions. *Organizational Behavior and Human Performance*, 19:267–310.

Keeney, R. and Raiffa, H. (1976). *Decisions with Multiple Objectives: Preferences and Value Tradeoffs*. Wiley and Sons, New York.

Medin, D. (1989). Concepts and conceptual structure. *American Psychologist*, 44(12):1468–1481.

Rosch, E. and Lloyd, B. (1978). *Cognition and Categorization*. Earlbaum Associates, Hillsdale, NJ.

Schank, R., Collins, G., and Hunter, L. (1986). Transcending inductive category formation in learning. *Behavioral and Brain Sciences*, 9(4):639–651.

Smith, E. and Medin, D. (1981). *Categories and concepts*. Havard University Press, Cambridge, MA.